# Xputer: Bridging Data Gaps with NMF, XGBoost, and a Streamlined GUI Experience

Saleena Younus[1,2,3], Lars Rönnstrand[1,2,3,4] and Julhash U. Kazi[1,2,3*]

[1]Division of Translational Cancer Research, Department of Laboratory Medicine, Lund University, Lund, Sweden.

[2]Lund Stem Cell Center, Department of Laboratory Medicine, Lund University, Lund, Sweden.

[3]Lund University Cancer Centre (LUCC), Lund University, Lund, Sweden.

[4]Department of Hematology, Oncology and Radiation Physics, Skåne University Hospital, Lund, Sweden

*Corresponding author: Email. Kazi.uddin@med.lu.se

**Abstract**

The rapid proliferation of data across diverse fields has accentuated the importance of accurate imputation for missing values. This task is crucial for ensuring data integrity and deriving meaningful insights. In response to this challenge, we present Xputer, a novel imputation tool that adeptly integrates Non-negative Matrix Factorization (NMF) with the predictive strengths of XGBoost. One of Xputer's standout features is its versatility: it supports zero imputation, enables hyperparameter optimization through Optuna, and allows users to define the number of iterations. For enhanced user experience and accessibility, we have equipped Xputer with an intuitive Graphical User Interface (GUI) ensuring ease of handling, even for those less familiar with computational tools. In performance benchmarks, Xputer not only rivals the computational speed of established tools such as IterativeImputer but also often outperforms them in terms of imputation accuracy. Furthermore, Xputer autonomously handles a diverse spectrum of data types, including categorical, continuous, and Boolean, eliminating the need for prior preprocessing. Given its blend of performance, flexibility, and user-friendly design, Xputer emerges as a state-of-the-art solution in the realm of data imputation.

## Introduction

The task of managing missing data points is an essential component of the preliminary data preparation phase, particularly in the context of data science applications such as machine learning models. A comprehensive dataset with no missing values is a prerequisite for a multitude of algorithms, emphasizing the necessity to address and rectify the absence of certain values. A rudimentary approach to resolve this issue might involve substituting these gaps with zeros or the mean of the column values. However, such methods could be an oversimplification, failing to take into account the inherent structural relationships within the data [1]. An advanced and well-accepted technique for managing missing data is the multiple imputation method, which was first proposed by Donald Rubin in 1978 [2]. This approach recognizes the uncertainty that comes with the imputation of missing values by generating multiple complete datasets. Each of these datasets offers a distinct but statistically plausible replacement for the missing data. The individual analysis results derived from these multiple datasets are then amalgamated to yield a single, more robust result. This process not only provides a method for filling in the missing data but also for estimating the statistical properties for subsequent analysis [3]

The Multivariate Imputation by Chained Equations (MICE) protocol, also referred to as Fully Conditional Specification (FCS) or Sequential Regression Multiple Imputation (SRMI), is a widely recognized imputation methodology with a process that shares similarities to multiple imputation. This technique is built on the premise of a round-robin approach, whereby every feature in the dataset that contains missing data is modeled as a function of the remaining features, and this process is performed iteratively [4, 5]. MICE is a flexible method that can handle various types of variables, making it a popular choice for dealing with missing data. Nevertheless, while MICE is a robust and versatile method, it is important to be cognizant of its potential limitations. Specifically, if MICE is implemented using its default settings, it may not fully capture non-linear relationships within the data during the imputation process. The default MICE process employs linear regression models for imputing missing values in continuous variables and logistic regression for binary and ordinal variables, both of which inherently assume a linear relationship between the predictors and the target variable [6, 7]. Consequently, these models may not be well-suited to data that exhibit non-linear relationships. There are ways to overcome this limitation. For instance, by integrating machine learning algorithms such as decision trees or random forests in the MICE process, non-linear relationships can be taken into account during imputation [8]. It is, therefore, crucial to understand the characteristics of the dataset at hand and to adjust the imputation strategy accordingly to ensure accurate results.

The application of tree-based models, such as those employing Random Forests or Classification and Regression Trees (CART), for data imputation presents a potentially powerful methodology [9, 10]. This is particularly relevant when data exhibit complex non-linear associations that could be otherwise challenging to handle with traditional linear methods. Tree-based models implement a hierarchical decision structure, which allows them to adeptly manage and exploit non-linear relationships embedded within the data. They thus offer an advanced tool for predicting missing data values under complex interaction scenarios. However, the adoption of such methodologies is not without its own set of challenges. Tree-based models, particularly those of a high complexity, can demand substantial computational resources, posing a hindrance for large-scale datasets [1]. Additionally, their performance can suffer in situations where a significant proportion of the data is missing. The difficulty of accurately predicting missing values increases as the amount of missing data rises, which may result in less reliable imputations. Furthermore, while tree-based models provide a robust method for capturing complex structures within data, the predictions they provide can be unstable due to their sensitivity to small changes in the training data, which can lead to substantial changes in the tree structure and subsequently in the imputed values. Thus, while tree-based models constitute a versatile tool for imputing missing data in complex scenarios, these limitations need to be carefully considered in the context of the specific dataset and research question at hand.

The process of imputing missing data should not be perceived as a rudimentary or minor phase in data preprocessing. Instead, it should be considered a critical procedure that can profoundly affect downstream analyses and the consequent outcomes. The decision regarding the choice of imputation technique should not be arbitrary or casual; instead, it should be guided by a careful assessment of the unique characteristics of the dataset and the specific requirements of the subsequent analysis [11]. The impacts that various imputation methods can have on the final results of the data analysis are not trivial and can be substantial. Thus, it is essential to assess the inherent assumptions accompanying each imputation method and to understand how they might influence the findings of the analysis. Notably, an inappropriate choice of an imputation method might introduce bias, distort relationships, or mask patterns in the data, leading to misleading or incorrect conclusions [12]. Therefore, understanding the properties and limitations of various imputation methods and their suitability to the specific data context is paramount for a robust and reliable data analysis [13].

In this article, we present a unique approach to missing value imputation that leverages the strengths of matrix factorization and the XGBoost algorithm. Our innovative method begins with an initial phase of imputation utilizing matrix factorization, which is a technique primarily aimed at reducing bias associated with the onset of imputation. Matrix factorization techniques help uncover the latent features underlying the interactions of the input data, thus providing a more robust and informed basis for initiating the imputation process. This step effectively minimizes the introduction of potential errors or biases at the very outset, thereby contributing to the overall accuracy and reliability of the imputation process. Following this, our approach utilizes XGBoost, a powerful gradient boosting framework, to predict missing values iteratively. In this phase, overfitting and underfitting - two common challenges in machine learning - are mitigated through automated hyperparameter optimization. The XGBoost model's hyperparameters are fine-tuned in a manner that balances bias and variance, thus preventing the model from becoming too complex (overfitting) or too simple (underfitting). To further enhance the precision of our imputation, we employ a process of iterative refinement of the missing value prediction. In essence, the imputation process is performed iteratively, with each iteration serving to improve the accuracy of the estimated missing values based on the results of the preceding iteration. Finally, in order to promote accessibility and ease of use, we have encapsulated this innovative imputation methodology within a user-friendly graphical user interface (GUI). This allows users, even those with limited programming or technical experience, to conveniently apply our imputation method to their own data. Thus, this powerful imputation tool can be broadly utilized across a range of disciplines and applications where missing data is a challenge.

**The architecture of Xputer**

To construct the foundation of Xputer, we combined NMF with a supervised learning model. While there is a plethora of supervised learning algorithms suitable for imputation, especially when considering prediction performance and computational efficiency, our choice to pair NMF with XGBoost was informed by a specific aim: addressing potential non-linear relationships in the unseen data. XGBoost is a powerful gradient boosting algorithm that can effectively handle both categorical and continuous data. It stands out for its speed when predicting large datasets and offers a high degree of flexibility in model tuning, which makes it an optimal choice for this integration [14]. However, it is worth noting that other gradient boosting frameworks, such as LightGBM, could serve as viable alternatives [15]. LightGBM also boasts efficient computational and memory usage, and it has shown itself to be an effective tool in predictive modelling scenarios. Thus, our strategy is designed to enhance imputation accuracy by leveraging the strengths of both NMF and XGBoost, thereby accommodating the diverse nature and complexity of data we might encounter.

The structural design of the Xputer is succinctly illustrated in Figure 1. This framework incorporates an array of interconnected modules, each serving a distinct functionality. Beginning with the preprocessing module (Figure 1A), this component undertakes the initial responsibility of data preparation. It executes tasks such as eliminating or correcting missing or erroneous data, standardizing numerical values, and transforming categorical variables into a format suitable for subsequent analyses. The subsequent step in the Xputer framework is the matrix factorization module (Figure 1B). This module leverages linear algebraic techniques to decompose a complex matrix into simpler, interpretable matrix constituents. This operation is pivotal to the simplification of high-dimensional data and the identification of inherent structure and relationships within the data. Next, the architecture encompasses the XGBoost imputation module (Figure 1C), a sophisticated component applying a gradient-boosting framework. This module plays an instrumental role in handling missing data, and extrapolating probable values based on the observed patterns in the existing data. It deploys a set of machine learning models, namely decision trees, which sequentially correct residuals of the preceding model, thereby enhancing the prediction accuracy. Finally, the iterative imputation module (Figure 1D) is incorporated into the Xputer design. This module cyclically estimates missing values, refining its predictions in each iteration until it reaches an optimal solution. This methodical process ensures that the imputation of missing data is not merely an isolated guess but a robust, data-informed estimation [16]. Together, these modules integrate to form the comprehensive architecture of Xputer, as visually encapsulated in Figure 1. Each module's role and interaction contribute to the efficacy and precision of the overall data analysis and prediction system.

*The preprocessing module:* Within this process, Xputer utilizes a Pandas Data Frame where sample identifiers are categorized in the index and feature designators are in the header. It commences by performing a scrutiny for absent or inoperative values, denoted by indicators such as 'NaN', 'NAN', 'Nan', 'nan', 'NA', '#NA', 'N/A', 'NA#', '#VALUE!', '#DIV/0!', etc., and proceeds to substitute any instances of these with the canonical 'np.nan'. In addition, it affords the user the option to convert zeros into 'np.nan' for later imputation. Subsequently, Xputer conducts an individual analysis for each column, distinguishing between continuous and categorical data points. If a specific type of data, say categorical, represents more than 60% of the column's contents, Xputer categorizes the entire column as such and replaces any numerical data with 'np.nan'. This protocol applies reciprocally to columns with over 60% numerical data, which are then considered as columns of continuous values and any non-numerical entries are replaced with 'np.nan'. Categorical data points are subsequently subject to encoding via a label encoder. In case a column fails to meet those criteria, leaves out of imputation. Post-processing, Xputer yields three separate Pandas DataFrames: a cleaned dataset, an encoded dataset, and a pre-imputed encoded dataset. The preprocessing module can function autonomously and be invoked through the "from xputer.utils import preprocessing_df" command.

*Adaptive matrix factorization:* Our adaptive matrix factorization strategy represents an advanced methodology designed to enhance the quality of pre-imputed data through the deployment of matrix factorization techniques. This process is characterized by its inherent adaptability, altering its computational approach based on the specific characteristics of the data at hand. In circumstances where the data set is devoid of negative values, the strategy deploys iterative sub-grouped non-negative matrix factorization (NMF). NMF is a technique that decomposes the original data into two non-negative matrix factors, providing an efficient and effective tool for representing the data and capturing the underlying structure [17]. As a consequence of NMF's non-negativity constraints, the resulting components are more interpretable and can offer a robust reconstruction of the original data. Conversely, in the presence of negative values within the dataset, the strategy resorts to Singular Value Decomposition (SVD). This method is a general matrix factorization technique that decomposes the data into three distinct matrices [18]. SVD is versatile and applicable to any real-valued matrix, providing an elegant solution when the non-negativity condition of NMF is not met. The adaptive matrix factorization strategy operates on both the encoded dataset and the pre-imputed encoded dataset, aiming to produce either a selectively transformed dataset—where only NaN values are replaced—or a fully reconstructed dataset depending on the specific requirements of the analysis. This dynamic approach ensures that the most suitable factorization method is applied, accounting for the unique attributes of the data and optimizing the precision of the imputation process.

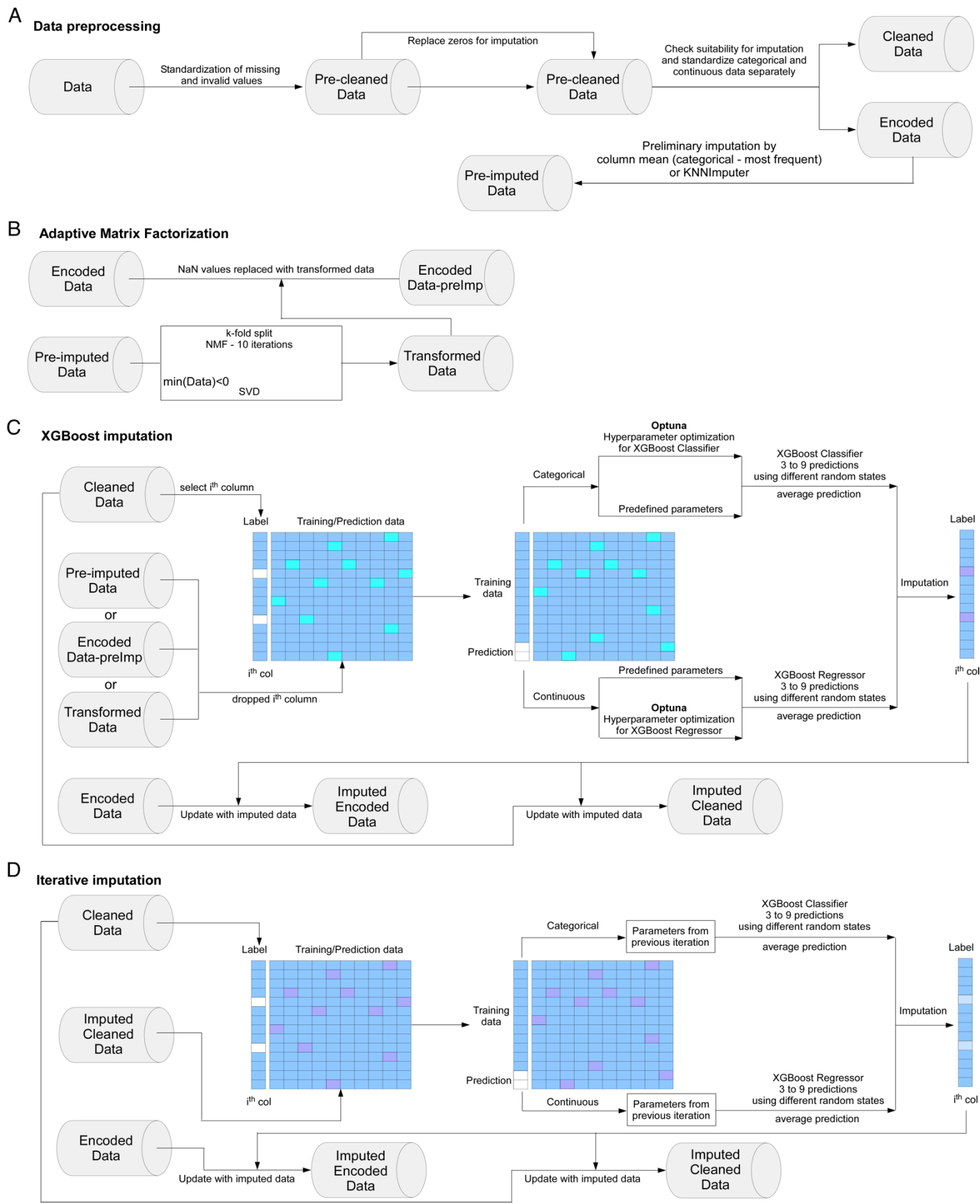

**Figure 1. The structure of Xputer.** The core structure of Xputer includes (A) data preprocessing unit, (B) adaptive matrix factorization unit, (C) XGBoost hyperparameter search and model implementation unit, and (D) iterative imputation unit.

*XGBoost imputation:* Xputer commences the imputation process by processing each column independently. It initiates this operation by selecting a column from the cleansed data, wherein categorical variables have not been encoded, and subsequently identifies the data type. Upon the determination of whether imputation is necessary for the specific column, the column in question is expunged from the pre-imputed or NMF/SVD-transformed data, thereby serving as a training or prediction set. Subsequent to the discernment of the data type, Xputer undertakes the decision as to whether a classifier or regressor is apt for the imputation task. For the optimization of the XGBoost hyperparameters, we have incorporated Optuna, a robust hyperparameter optimization framework [19]. However, prior to commencing the hyperparameter search, Xputer evaluates the number of samples, necessitating more than 50, and validates the sample-to-feature ratio, mandating a minimum ratio of 4. Furthermore, it reviews the number of features, with a maximum allowance of 100 columns with missing values. Given these conditions, Xputer determines if the hyperparameter search needs to be executed, or if the deployment of predefined parameters is more pertinent. Regardless of the path chosen, Xputer employs a 70% subsampling strategy. In the event a hyperparameter search is performed, the system is meticulously designed to evade both overfitting and underfitting of the model. Once the hyperparameters are solidified, Xputer executes a minimum of three predictions (with an option to select up to nine) and employs an average value for the imputation. Both the cleansed data and encoded data are consequently updated with the predicted values. Additionally, Xputer accumulates the XGBoost parameters for each column from this phase for potential utilization in iterative imputation procedures in the future.

*Iterative imputation:* Xputer has been architecturally devised to execute an iterative process of imputation. In this scheme, values that have been previously estimated and subsequently imputed are employed for subsequent predictions, effectively utilizing iterative refinement for the missing data estimation. This mechanism proves particularly efficacious when dealing with datasets afflicted by a substantial proportion of missing values. In such scenarios, every column is sequentially updated with novel predictions, each one informed by the newly imputed values from the previous round. This creates a feedback loop where each iteration enhances the accuracy of imputations in the next, thus leveraging the potential for continuous improvement in the estimation of missing data. As of now, Xputer accommodates a range of iterations from a minimum of one to a maximum of nine, offering flexibility in choosing the depth of refinement according to the specifics of the dataset and the imputation requirements. This iterative imputation process propagates information from observed to unobserved values, making it a powerful method to reconstruct missing data in large datasets.

## Results

**Supervised learning algorithms hold the superior predictive capacity for missing value imputation.**

In the initial phase of our study, a comparative analysis was conducted among several supervised and unsupervised machine learning algorithms to evaluate their efficacy in the imputation of missing values. The chosen dataset for this investigation comprised of a heterogeneous collection of gene expression data, including both microarray and RNAseq data, covering 400 genes (features). To facilitate this comparison, we artificially introduced missing values into the dataset by removing data ranging from 1% up to 25%, thereby generating 25 unique datasets. An initial imputation was then carried out using column mean values, which served as a foundation for the subsequent application of 27 distinct regressors, each tasked with predicting the missing values in every column (Figure 2A). The array of machine learning algorithms under examination comprised unsupervised methods such as autoencoders, non-negative matrix factorization (NMF), and principal component analysis (PCA), which were juxtaposed against their supervised counterparts. The performance of these algorithms was compared based on the log2-transformed mean squared error (MSE) and the log2-transformed computational time in seconds. The majority of supervised algorithms exhibited comparable performances, with the exceptions of RANSAC and the Gaussian process which differed (Figure 2B). In terms of time efficiency, PCA emerged as the most proficient (Figure 2C); however, its performance lagged when evaluated based on the NegLog2RMSL (Figure 2D-E). The comparative results suggested that supervised algorithms are relatively more efficient in the imputation process than most unsupervised algorithms, with NMF being the exception. However, it should be emphasized that none of the algorithms were optimized for this specific task. Therefore, it is possible that the performance of some algorithms could improve if optimization were to be implemented.

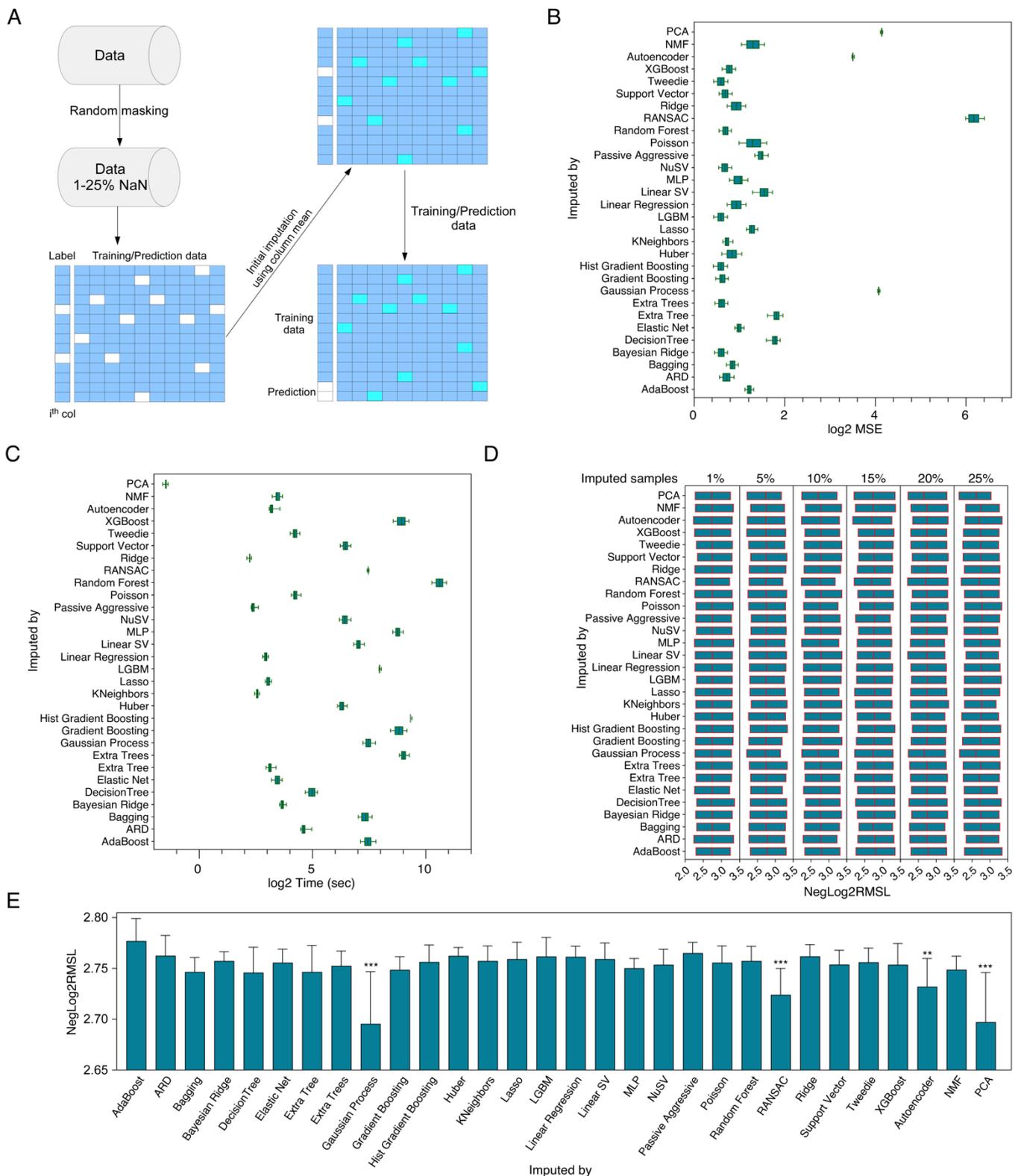

**Figure 2. Comparative evaluation of various algorithms' functionality.** (A) A composite dataset consisting of RNAseq and microarray data, dimensioned at 1770 X 400, underwent random masking to instigate missing values, ranging from 1% to 25%, thus creating 25 distinct data matrices. In the case of unsupervised learning algorithms, namely autoencoder, NMF, and PCA, the imputation of the entire dataset was accomplished via the column mean through SimpleImputer prior to the execution of the algorithm. For supervised algorithms, an initial identification of columns with NaNs was carried out. An individual column containing NaN was isolated and designated as the label, while the remainder of the data underwent imputation via the column mean with the aid of SimpleImputer. Following the preliminary imputation, the data was segregated into training and predictive datasets corresponding to non-NaN and NaN rows in the label column. (B) Subsequent to imputation with a specific algorithm, imputed values underwent a comparative analysis with their original counterparts, facilitating the determination of the mean squared error. Each box demonstrates 25 measurements employing missing values varying from 1%-25% (in increments of 1%), while the bar represents the minimum to maximum values range. (C) The duration necessitated to carry out imputation utilizing a specific

algorithm. (D-F) Six imputed datasets were employed to ascertain the prediction performance of the Trametinib response via the XGBoost algorithm.

**Xputer hyperparameters testing**

In assessing the performance of Xputer, hyperparameter optimization, though optional, can potentially enhance its efficacy. Six diverse datasets facilitated an exploration of Xputer hyperparameters' impact on imputation. We artificially introduced missing values at varying percentages (10% to 40%) across both categorical and continuous data types within each dataset. Our initial inquiry centered on the influence of pre-imputation on the ultimate imputation outcomes. We employed three distinct strategies: *ColumnMean:* missing continuous values were substituted with the respective column mean, whereas missing categorical or Boolean entries were replaced by the most frequently occurring value, *KNNImputer:* utilized the sklearn's KNNImputer to address NaNs across all columns, and *MixType:* continuous NaNs were supplanted by the column mean, while the remaining were managed using KNNImputer. For the majority of columns across the datasets, there was no discernible preference for any particular strategy. However, certain scenarios evidenced a tilt towards ColumnMean, KNNImputer, or MixType (Figure 3A). By default, Xputer constructs three XGBoost models for imputation, which can be expanded up to nine, leveraging the mean value for imputation. We juxtaposed the efficacy of utilizing 3, 6, or 9 models, maintaining other default parameters. Some attributes benefited from a larger model pool, yet the overall enhancement remained marginal (Figure 3B), implying that three models might suffice for robust imputation. The application of either NMF NaN-transformed or fully transformed data could potentially ameliorate imputation, especially in the presence of outliers (Figure 3C). Though our datasets did not manifest tangible gains from Optuna-driven hyperparameter tuning compared to predefined parameters (Figure 3D), we cannot preclude its potential benefits in other datasets. Notably, a reduced iteration count in Optuna consistently underperformed compared to default parameters. Thus, it's advised to either adhere to default settings or, if using Optuna, opt for a greater number of iterations. Finally, while augmenting the iteration count can confer minor enhancements, the default setting generally sufficed (Figure 3E).

**Dependency of Xputer's Imputation Time on Feature Quantity**

The imputation latency experienced with the Xputer system is intricately linked to the quantum of features necessitating imputation. Mechanistically, Xputer operates by scanning for null values within each column, subsequently embarking on a column-wise imputation procedure leveraging the XGBoost algorithm. Specifically, when a column is targeted for imputation, extant values therein are harnessed to construct the XGBoost predictive model, which then projects or fills the absent values. This modus operandi elucidates why columns teeming with null values considerably amplify the temporal demands of the imputation process (Figure 4A). Further analyses revealed a negative logarithmic relationship between the magnitude of missing values and the time commitment per unit of such values (Figure 4B). Intriguingly, while the aggregate sample volume appeared to exert negligible impact on imputation duration (Figure 4C), there was a robust correlation between an escalating feature count and the time invested per feature during imputation (Figure 4D). Comparative assessments with the IterativeImputer showcased that, under default configurations, IterativeImputer demonstrated superior speed for a diminutive feature set but followed Xputer with an expansive feature count. Moreover, when configured with a Random Forest estimator, the IterativeImputer consistently underperformed relative to Xputer, irrespective of the feature quantity (Figure 4E). In aggregate, the empirical evidence posits that Xputer's performance might decelerate as the feature count swells.

**Comparing Xputer with IterativeImputer**

In this analysis, we juxtaposed the imputation efficacy of Xputer against that of IterativeImputer. Notably, Xputer is intrinsically engineered to seamlessly manage a diverse range of data types, including categorical, bimodal, and continuous datasets. However, to ensure an equitable benchmark, particularly mirroring the primary function of IterativeImputer, our evaluation was confined to continuous data. The rigorousness of our evaluation was upheld through the employment of the Root Mean Square Error (RMSE) metric coupled with detailed density plots. This assessment spanned across scenarios with 10% to 40% missing data values in contrast to the original dataset. Our empirical findings revealed that Xputer demonstrated consistent and robust performance (Figure 5A). Conversely, the IterativeImputer, under its default configuration, occasionally faltered. Yet, an augmentation of the IterativeImputer with the Random Forest Regressor dramatically enhanced its imputation efficiency. This enhancement was further corroborated by the density plots, which unveiled a striking resemblance between the original dataset, Xputer-imputed data, and data imputed using the Random Forest Regressor combined with IterativeImputer (Figure 5B). Our comparative study underscores the inherent strengths of Xputer in data imputation. However, the synergy of the Random Forest Regressor with IterativeImputer offers a commendable alternative, illustrating the potential of hybrid methodologies in advancing data imputation techniques.

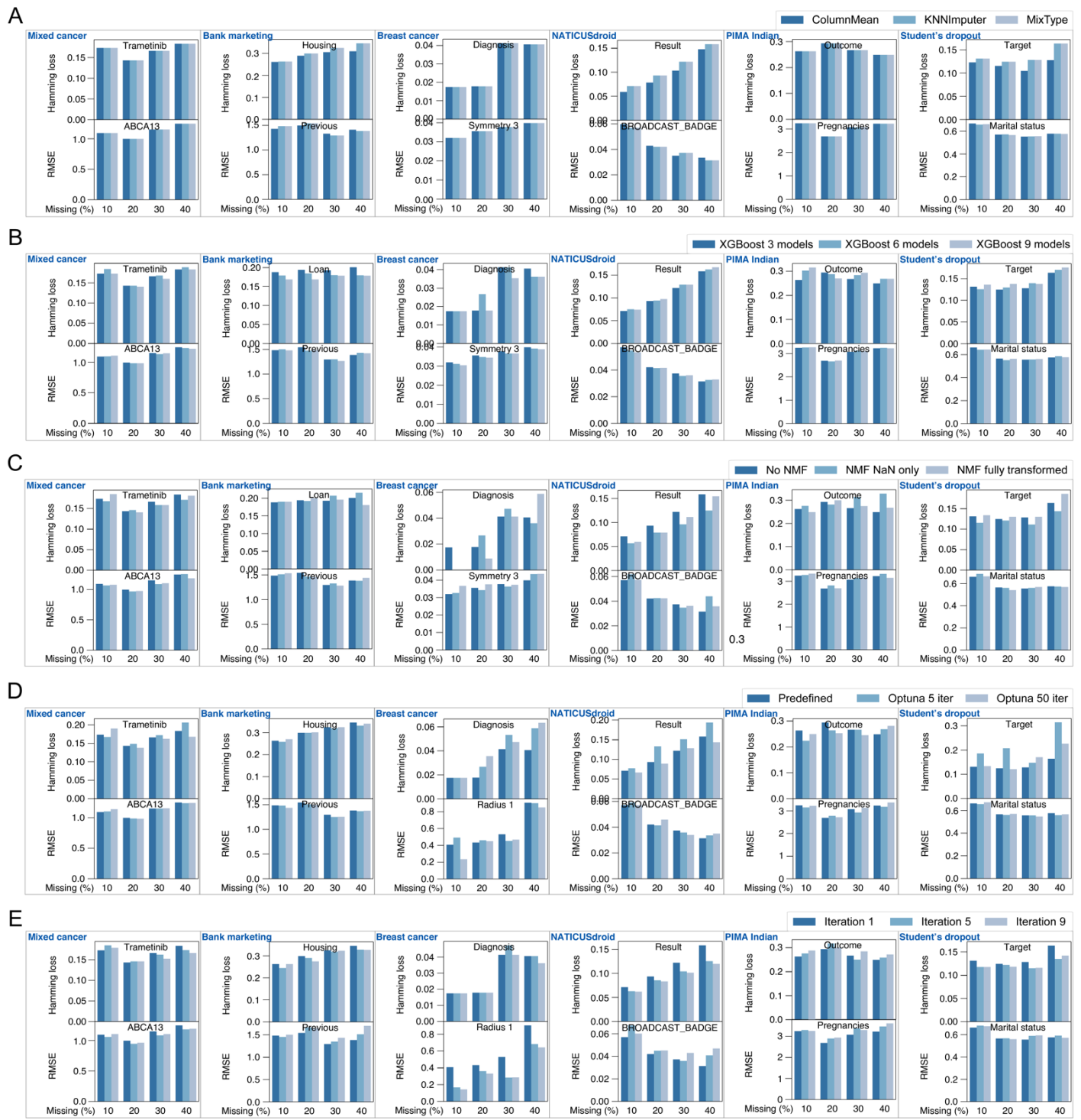

**Figure 3. Xputer hyperparameter evaluation.** For a comprehensive assessment of Xputer's individual hyperparameters, six distinct datasets were employed, each infused with a diverse spectrum of artificially introduced missing values. These modified datasets were subsequently juxtaposed against their original values. The hyperparameters under examination encompassed (A) pre-imputation strategies, (B) the count of XGBoost models for ensembling, (C) matrix factorization techniques, (D) XGBoost hyperparameter optimization using Optuna, and (E) the total iteration count.

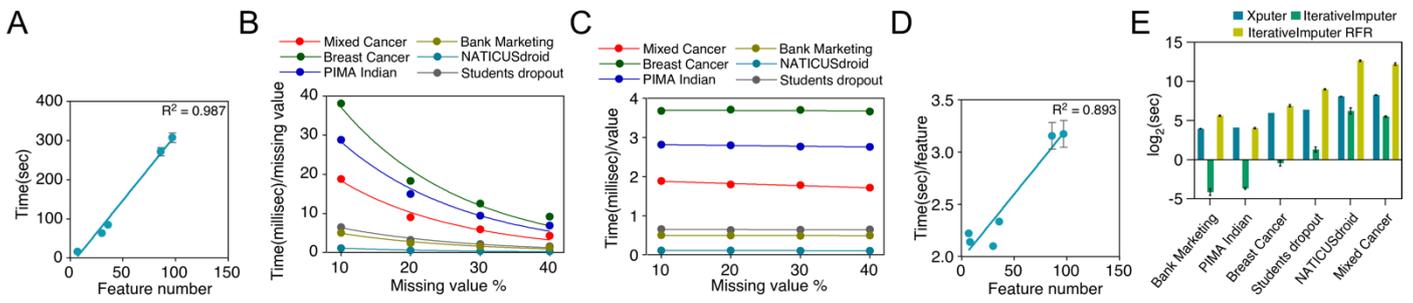

**Figure 4. Assessment of Imputation Duration.** In this analysis, we utilized continuous data encompassing 7 to 97 features and sample sizes varying from 500 to 30,000. This evaluation covered various metrics: (A) time taken for imputation relative to the number of features, (B) time per unit of missing value, (C) time per individual value, and (D) imputation time for each feature in relation to the total number of features. Additionally, (E) we compared our results with the IterativeImputer's performance.

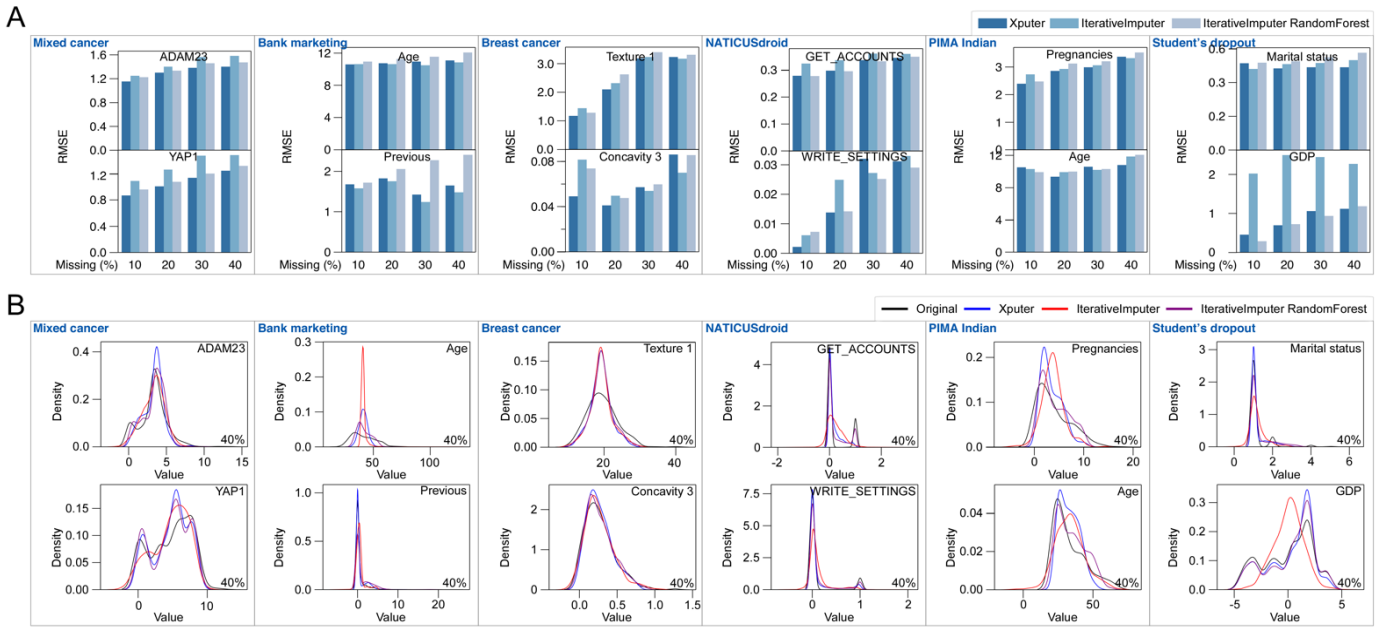

**Figure 5. Comparative Analysis of Xputer and IterativeImputer.** Using six continuous-data datasets, we assessed the performance of Xputer against IterativeImputer. Performance comparisons were conducted employing both the RMSE metric (A) and density plots (B) across varying percentages of missing values.

## Discussion

Addressing the challenge of missing value imputation is paramount across a range of disciplines [20-22]. In bioinformatics, dealing with the multifaceted and distinct characteristics of various data types including RNAseq, proteomic and metabolomic data poses unique challenges for missing value imputations [20]. In the world of marketing, complete datasets are essential to derive accurate consumer insights, optimize advertising strategies, and predict market trends [23]. Missing values can lead to misleading segmentation or incomplete customer profiles, which can adversely impact campaign efficacy and return of investment (ROI). Similarly, in the banking and financial sectors, accurate data is the backbone of risk assessments, credit scoring, and investment decisions. Missing or incorrect data can lead to misguided financial strategies, erroneous lending decisions, or misestimation of market risks [24]. Household surveys, often used for demographic research or to assess consumer behavior, are another domain where imputation plays a pivotal role. Such datasets often contain gaps due to non-responses or incomplete survey returns, and imputation is vital to provide a comprehensive view of the population being studied [25]. Given the increasing reliance on data across industries, the need for sophisticated and accurate imputation methods becomes ever more crucial. The heterogeneous nature of datasets across sectors underscores the need for robust and tailored imputation methods to ensure data integrity, accurate predictions, and informed decision-making.

Given the complexities highlighted above, we have developed Xputer that combines flexibility, speed and diversity. In our comprehensive analysis, the role of supervised learning algorithms in enhancing the precision of imputation emerged unmistakably [26]. Through rigorous evaluation using datasets artificially populated with missing values, our findings affirmed the superior predictive capacity of supervised algorithms over their unsupervised counterparts for the imputation of missing values. This advantage held true despite the fact that none of the evaluated algorithms was specifically optimized for the imputation task, suggesting innate strengths within supervised learning paradigms. The only unsupervised exception, NMF, hints at potential niches wherein unsupervised learning could have competitive or even advantageous roles. Furthermore, the detailed hyperparameter testing of Xputer showcased its potential versatility. While it was evident that hyperparameter tuning can yield marginal improvements, for many datasets, the default configurations of Xputer often delivered satisfactory imputation outcomes. Notably, the nature of missing data, whether categorical or continuous, and its percentage appeared to influence the preferred imputation strategy. The relationship between Xputer's imputation time and the volume of features underscores a salient operational aspect of the software. Xputer's meticulous column-wise approach, relying on XGBoost, illuminates the escalating temporal requirements with an increasing proportion of missing values. The observed logarithmic relationship between time and magnitude of missing values points to diminishing returns in efficiency as missing data accrue. Interestingly, this observation is juxtaposed with the rather consistent temporal demands across varying sample sizes, revealing feature count as the more prominent determinant of imputation latency.

The comparative evaluation with IterativeImputer, a similar version of MICE [5] offered insights into the nuances of these two methodologies. While Xputer's intrinsic design accommodates a diverse gamut of data types, IterativeImputer requires specific configurations, like the inclusion of the Random Forest Regressor, to reach peak performance, particularly in the context of continuous data. The observational consistency between the original and imputed datasets, especially with Xputer and the optimized IterativeImputer, elucidates the potential of these tools in preserving the inherent structure and nuances of gene expression datasets. In culmination, our exploration underscores the importance of algorithmic choice and optimization in the imputation of missing gene expression values. While supervised algorithms, embodied by Xputer, typically excel in these endeavors, the context-specific merits of unsupervised algorithms, or hybrid solutions like IterativeImputer combined with Random Forest, cannot be overlooked. The findings also emphasize the necessity to understand the operational nuances of each tool, ensuring that researchers can harness them effectively, optimizing both imputation accuracy and computational efficiency.

## ACKNOWLEDGMENTS

We thank the Lund University Library for their generous assistance in covering the costs associated with open access publication.

## FUNDINGS

This research was supported by the the Crafoord Foundation (J.U.K # 20230775), the Swedish Cancer Society (J.U.K. #19 0004 FE and L.R. #21 1444 Pj), the Swedish Research Council (L.R. #2021-03055), the Swedish Childhood Cancer Foundation (J.U.K. # PR2022-0106), and SUS Stiftelser och Donationer (L.R. #95512), Governmental Funding of Clinical Research within the National Health Service (ALF) (L.R. # 40609).

## DATA AVAILABILITY

The Python scripts are designed for those with basic computer skills and a foundational understanding of handling spreadsheet data, similar to what's done in Microsoft Excel. To further simplify the user experience, we offer a Graphical User Interface (GUI). The Python library, xputer, can be downloaded from the Python Package Index (PyPI) at www.pypi.org/project/xputer/ or from our GitHub repository at www.github.com/kazilab/xputer. All associated data is openly accessible.

## DATA PROCUREMENT

The "Mixed cancer" data was from the BeatAML and scDEAL studies [27, 28]. The "Bank marketing" data was from archive.ics.uci.edu/dataset/222/bank+marketing, the "Breast cancer" data was collected from archive.ics.uci.edu/dataset/17/breast+cancer+wisconsin+diagnostic, the "NATICUSdroid" data was from archive.ics.uci.edu/dataset/722/naticusdroid+android+permissions+dataset, the "PIMA Indian" data was downloaded from www.kaggle.com/datasets/uciml/pima-indians-diabetes-database, and the "Student's dropout" data was collected from archive.ics.uci.edu/dataset/697/predict+students+dropout+and+academic+success.

**Reference**

1. Hastie T, Tibshirani R, Friedman J. The Elements of Statistical Learning: Data Mining, Inference, and Prediction. Springer, 2016.
2. Rubin DB. Multiple Imputation for Nonresponse in Surveys. John Wiley & Sons, Inc., 1987.
3. Little RJ, Rubin DB. Statistical Analysis with Missing Data. John Wiley & Sons, Inc., 2019.
4. Azur MJ, Stuart EA, Frangakis C et al. Multiple imputation by chained equations: what is it and how does it work?, Int J Methods Psychiatr Res 2011;20:40-49.
5. van Buuren S, Groothuis-Oudshoorn K. mice: Multivariate imputation by chained equations in R, Journal of statistical software 2011;45:1-67.
6. Raghunathan TE, Lepkowski JM, Hoewyk JV et al. A Multivariate Technique for Multiply Imputing Missing Values Using a Sequence of Regression Models, Survey Methodology 2001;27:85-95.
7. Su YS, Gelman A, Hill J et al. Multiple Imputation with Diagnostics (mi) in R: Opening Windows into the Black Box, Journal of statistical software 2011;45:1-31.
8. Shah AD, Bartlett JW, Carpenter J et al. Comparison of random forest and parametric imputation models for imputing missing data using MICE: a CALIBER study, Am J Epidemiol 2014;179:764-774.
9. Trendowicz A, Jeffery J. Classification and Regression Trees. Software Project Effort Estimation. Springer, Cham, 2013, 295–304.
10. Breiman L. Random Forests, Machine Learning 2001;45:5-32.
11. Donders AR, van der Heijden GJ, Stijnen T et al. Review: a gentle introduction to imputation of missing values, J Clin Epidemiol 2006;59:1087-1091.
12. Sterne JA, White IR, Carlin JB et al. Multiple imputation for missing data in epidemiological and clinical research: potential and pitfalls, BMJ 2009;338:b2393.
13. van Buuren S. Flexible Imputation of Missing Data. Chapman and Hall/CRC, 2018.
14. Chen T, Guestrin C. XGBoost: A Scalable Tree Boosting System, KDD '16: Proceedings of the 22nd ACM SIGKDD International Conference on Knowledge Discovery and Data Mining 2016:785–794.
15. Ke G, Meng Q, Finley T et al. LightGBM: A Highly Efficient Gradient Boosting Decision Tree, Advances in Neural Information Processing Systems 2017;3149-3157.
16. Dong Y, Peng CY. Principled missing data methods for researchers, Springerplus 2013;2:222.
17. Lee DD, Seung HS. Learning the parts of objects by non-negative matrix factorization, Nature 1999;401:788-791.
18. DeSa RJ, Matheson IB. A practical approach to interpretation of singular value decomposition results, Methods Enzymol 2004;384:1-8.
19. T. A, S. S, T. Y et al. Optuna: A Next-generation Hyperparameter Optimization Framework, Proceedings of the 25th {ACM} {SIGKDD} International Conference on Knowledge Discovery and Data Mining 2019;KDD 19:2623–2631.
20. Vanderaa C, Gatto L. Revisiting the Thorny Issue of Missing Values in Single-Cell Proteomics, J Proteome Res 2023.
21. Choi J, Lim KJ, Ji B. Robust imputation method with context-aware voting ensemble model for management of water-quality data, Water Res 2023;243:120369.
22. Mukherjee K, Gunsoy NB, Kristy RM et al. Handling Missing Data in Health Economics and Outcomes Research (HEOR): A Systematic Review and Practical Recommendations, Pharmacoeconomics 2023.
23. Anand V, Mamidi V. Multiple Imputation of Missing Data in Marketing. 2020, 1-6.
24. Fujimoto S, Mizuno T, Ishikawa A. Interpolation of non-random missing values in financial statements' big data using CatBoost, Journal of Computational Social Science 2022;5:1281-1301.
25. Mirzaei A, Carter SR, Patanwala AE et al. Missing data in surveys: Key concepts, approaches, and applications, Research in Social and Administrative Pharmacy 2022;18:2308-2316.
26. Getz K, Hubbard RA, Linn KA. Performance of Multiple Imputation Using Modern Machine Learning Methods in Electronic Health Records Data, Epidemiology 2023;34:206-215.
27. Chen J, Wang X, Ma A et al. Deep transfer learning of cancer drug responses by integrating bulk and single-cell RNA-seq data, Nat Commun 2022;13:6494.
28. Bottomly D, Long N, Schultz AR et al. Integrative analysis of drug response and clinical outcome in acute myeloid leukemia, Cancer Cell 2022;40:850-864 e859.

# Xputer: Bridging Data Gaps with NMF, XGBoost, and a Streamlined GUI Experience

## Supplementary methods

Procedure for configuring the Xputer within a Python-based environment:

**Installation**

1. *Install Anaconda:* Download the latest version of Anaconda from www.anaconda.com. Install by double-clicking the downloaded installer. When the installer asks for the installation type, choose "Just Me".
2. *a) To Install Xputer on Windows*:
   - Locate the "Anaconda PowerShell Prompt" using the Windows search function.
   - Open it and input the following command.
   - `pip install xputer`
   - Hit "Enter" to begin the installation of Xputer and its associated packages.

   *b) To Install Xputer on MacOS*

   - Locate the "Terminal" application and open it.
   - Prior to installing the package, initiate the base conda environment.
   - Do this by entering the command below and then pressing "Enter":
   - `conda activate`
   - "base" should be prepended.
   - Type the following command and press "Enter" to install Xputer:
   - `pip install xputer`
   - To deactivate the conda environment use the following command:
   - `conda deactivate`

   *C) Platform-independent alternative*

   - Launch Anaconda Navigator
   - From the navigator launch JupyterLab. It will open a browser window.
   - Click on Terminal from the browser window.
   - Type the following command in Terminal/PowerShell window:
   - `pip install xputer`
   - Press "Enter" to install Xputer.

**Run Xputer**

1. *Run on Windows:*
   - Locate the "Anaconda PowerShell Prompt" using the Windows search function.
   - Open it and input the following command.
   - `python -m xputer`
   - Press "Enter" to open the Xputer GUI

2. *Run on MacOS:*
   - Locate the "Terminal" application and open it.
   - Prior to installing the package, initiate the base conda environment.
   - Do this by entering the command below and then pressing "Enter":
   - `conda activate`
   - "base" should be prepended.
   - To run Xputer, type the following command and then press "Enter":
   - `python -m xputer`
   - To deactivate the conda environment use the following command:
   - `conda deactivate`

3. *To run on platform-independent alternative*
    - Launch Anaconda Navigator
    - From the navigator launch JupyterLab. It will open a browser window.
    - Click on Terminal from the browser window.
    - Type the following command in Terminal/PowerShell window:
    - `python -m xputer`
    - Press "Enter" to run Xputer GUI.

**Use Xputer from the GUI**

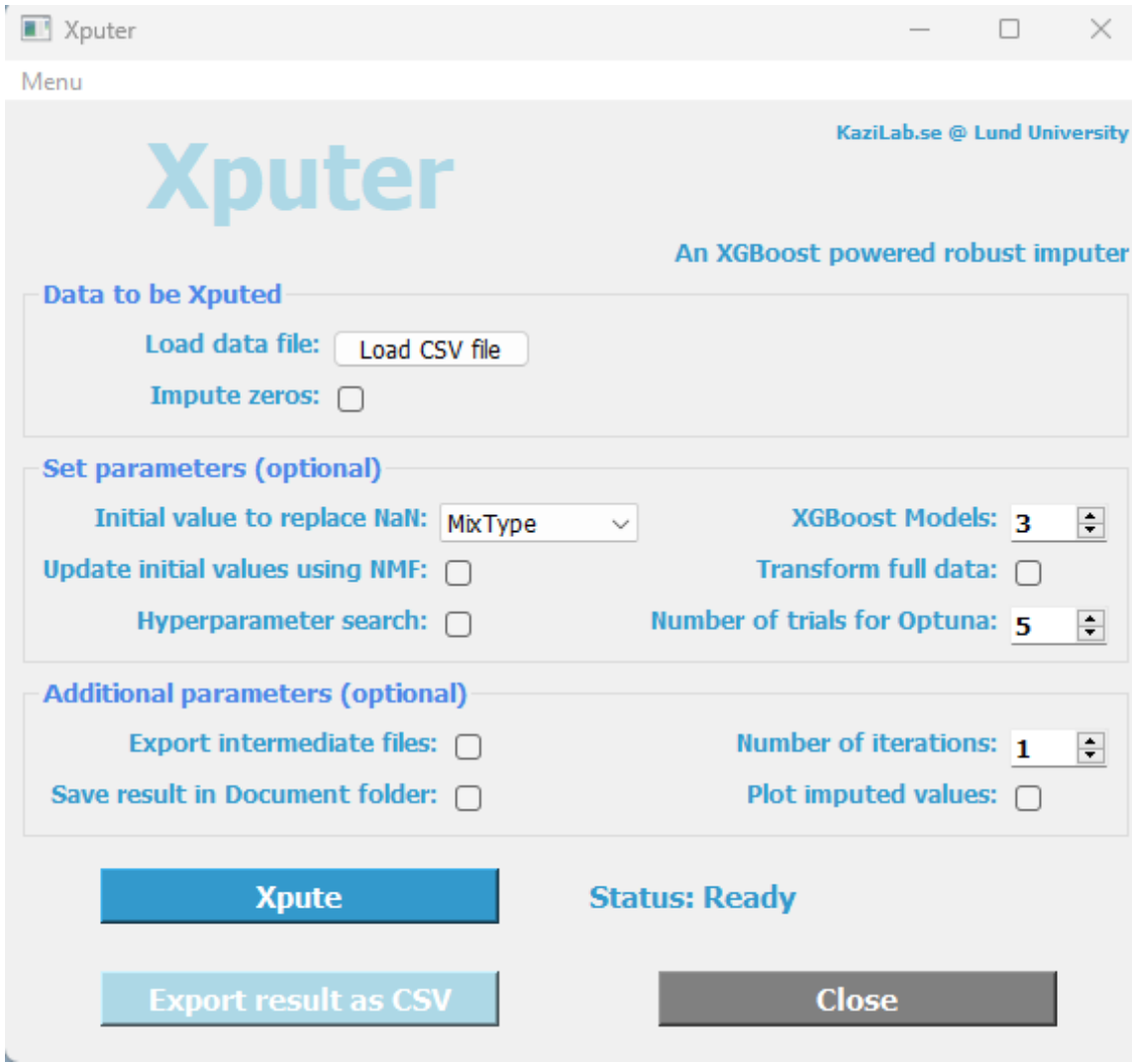

1. For imputation to proceed, the data should be supplied in a CSV format. In this format, the first column should list the sample names, while the top row should enumerate the feature names. See the example below:

| Sample | Feature 1 | Feature 2 | Feature 3 | Feature 4 | Feature 5 | Feature 6 | Feature 7 | Feature 8 |
|--------|-----------|-----------|-----------|-----------|-----------|-----------|-----------|-----------|
| Sample 1 | Data | Data | Data | Data | Data | Data | Data | Data |
| Sample 2 | Data | Data | Data | Data | Data | Data | Data | Data |
| Sample 3 | Data | Data | Data | Data | Data | Data | Data | Data |
| Sample 4 | Data | Data | Data | Data | Data | Data | Data | Data |

2. To load the data file in CSV format, click on "Load CSV file", locate the file, and press open.
3. Mouse over parameters to get inline help.
4. Click on "Xpute" to run imputation.

Optional options:

1. To impute zeros, select the "Impute zeros" checkbox.
2. You can replace initial NaN values using one of three methods from the dropdown list.
3. The "XGBoost Models" option, used for averaging the final imputation, allows a range from 3 to 9 models.
4. If "Update initial values using NMF" is selected, the data matrix will be factorized and reconstructed using either NMF or SVD (chosen if a negative value is present). Only the NaN values will then be replaced by the transformed values.
5. If "Transform full data" is selected, the data matrix will be factorized and reconstructed using either NMF or SVD (if a negative value is present). The fully transformed data will then be utilized for XGBoost prediction.
6. Checking "Hyperparameter search" will engage Optuna for XGBoost hyperparameter optimization. The "Number of trials for Optuna" setting determines the number of trials for this search.
7. If "Export intermediate files" is checked, Xputer will save all intermediate files to the "Document" folder.
8. "Number of iterations" lets you adjust the count of XGBoost iterations.
9. If "Save result in Document folder" is checked, results will be saved automatically in that folder.
10. Selecting "Plot imputed values" will generate and save both density plots and dot plots.

**Advanced options:**

To use in the pipeline:

```
from xputer import xpute

xpute function returns the imputed dataframe

imputed_df = xpute(df,
                  impute_zeros=False,
                  pre_imputation='MixType',
                  xgb_models=3,
                  mf_for_xgb=False,
                  use_transformed_df=False,
                  optuna_for_xgb=False,
                  optuna_n_trials=5,
                  n_iterations=1,
                  save_imputed_df=False,
                  save_plots=False,
                  test_mode=False)
```

or
```
from xputer import Xpute
xpute = Xpute(impute_zeros=False,
              pre_imputation='MixType',
              xgb_models=3,
              mf_for_xgb=False,
              use_transformed_df=False,
              optuna_for_xgb=False,
              optuna_n_trials=5,
              n_iterations=1,
              save_imputed_df=False,
              save_plots=False,
              test_mode=False)
imputed_df = xpute.fit(df)
```

df: pandas DataFrame with index (sample names) and header (features).
impute_zeros: Boolean.
pre_imputation: 'MixType', 'ColumnMean', 'KNNImputer'.
xgb_models: Integer, between 3 and 9.
mf_for_xgb: Boolean.
use_transformed_df: Boolean.
optuna_for_xgb: Boolean.
optuna_n_trials: Integer, between 5 and 50.
n_iterations: Integer, between 1 and 9.
save_imputed_df: Boolean.
save_plots: Boolean.
test_mode: Boolean.

**Use independent modules:**

```
from xputer import preprocessing_df
preprocessing_df  returns a clean_df, an encoded_df, and a preimputed_df
clean_df, encoded_df, preimputed_df = preprocessing_df(df,
                                        impute_zeros=False,
                                        pre_imputation='MixType',
                                        test_mode=False)
```

from xputer import cnmf

cnmf function returns two dataframes, NaN_imputed_by_NMF and Fully_transformed_df, when provided encoded_df and preimputed_df from preprocessing_df.

NaN_imputed_by_NMF, Fully_transformed_df = cnmf(encoded_df, preimputed_df)

from xputer import run_svd

run_svd function is similar to cnmf except here it initiates SVD instead of NMF.

NaN_imputed_by_svd, Fully_transformed_df = run_svd(encoded_df, preimputed_df)